\documentclass[letterpaper, 10 pt, conference]{ieeeconf}  % Comment this line out if you need a4paper

\IEEEoverridecommandlockouts                              % This command is only needed if 
\usepackage{graphics} % for pdf, bitmapped graphics files
\usepackage{graphicx}
\usepackage{mathptmx} % assumes new font selection scheme installed
\usepackage{amsmath} % assumes amsmath package installed
\usepackage{amssymb}  % assumes amsmath package installed
\usepackage{multirow}
\usepackage{algorithm}
\usepackage{newfloat}
\usepackage{listings}
\usepackage{subcaption}
\usepackage{xcolor}
\usepackage{algorithmic}
\usepackage{booktabs}
\usepackage{nameref}
\definecolor{mygreen}{HTML}{1b9e77}
\definecolor{myorange}{HTML}{d95f02}
\definecolor{myblue}{HTML}{4139bd} %534bbd
\definecolor{mymagenta}{HTML}{D62680} %e7298a
\definecolor{codegreen}{rgb}{0.0,0.6,0.0}

\title{\LARGE \bf
Ego-Motion Aware Target Prediction Module for Robust Multi-Object Tracking
}

\author{Navid Mahdian$^{1\dagger}$, Mohammad Jani$^{1\dagger}$, Amir M. Soufi Enayati$^{2}$, and Homayoun Najjaran$^{1,2}$% <-this % stops a space
\thanks{$^*\,$This work was supported by MITACS under the Accelerate Program Internship No. IT31385 and Sentire.}
\thanks{$^{1}\,$Department of Electrical and Computer Engineering, University of Victoria, Victoria, BC, Canada {\tt\small\{navidmahdian, mhmdjani, najjaran\}@uvic.ca}}
\thanks{$^{2}\,$Department of Mechanical Engineering, University of Victoria, Victoria, BC, Canada {\tt\small amsoufi@uvic.ca}}
\thanks{$\dagger$~Equivalent contribution as the common first authors.}
}

\begin{document}

\maketitle
\thispagestyle{empty}
\pagestyle{empty}

%%%%%%%%%%%%%%%%%%%%%%%%%%%%%%%%%%%%%%%%%%%%%%%%%%%%%%%%%%%%%%%%%%%%%%%%%%%%%%%%
\begin{abstract}

Multi-object tracking (MOT) is a prominent task in computer vision with application in autonomous driving, responsible for the simultaneous tracking of multiple object trajectories.
Detection-based multi-object tracking (DBT) algorithms detect objects using an independent object detector and predict the imminent location of each target.
Conventional prediction methods in DBT utilize Kalman Filter (KF) to extrapolate the target location in the upcoming frames by supposing a constant velocity motion model.
These methods are especially hindered in autonomous driving applications due to dramatic camera motion or unavailable detections.
Such limitations lead to tracking failures manifested by numerous identity switches and disrupted trajectories.
In this paper, we introduce a novel KF-based prediction module called the \textbf{E}go-\textbf{M}otion \textbf{A}ware Target \textbf{P}rediction (EMAP) module by focusing on the integration of camera motion and depth information with object motion models.
Our proposed method decouples the impact of camera rotational and translational velocity from the object trajectories by reformulating the Kalman Filter.
This reformulation enables us to reject the disturbances caused by camera motion and maximizes the reliability of the object motion model.
We integrate our module with four state-of-the-art base MOT algorithms, namely OC-SORT, Deep OC-SORT, ByteTrack, and BoT-SORT.
In particular, our evaluation on the KITTI MOT dataset demonstrates that EMAP remarkably drops the number of identity switches (IDSW) of OC-SORT and Deep OC-SORT by 73\% and 21\%, respectively. At the same time, it elevates other performance metrics such as HOTA by more than 5\%. Our source code is available at \texttt{https://github.com/noyzzz/EMAP}.

\end{abstract}

%%%%%%%%%%%%%%%%%%%%%%%%%%%%%%%%%%%%%%%%%%%%%%%%%%%%%%%%%%%%%%%%%%%%%%%%%%%%%%%%
\section{Introduction} \label{intro}
Multi-object tracking stands out as a pivotal task in computer vision, finding diverse applications, ranging from monitoring crowds via surveillance cameras to analyzing the intricate movements of soccer players on the field to tracking vehicles on the road and following marine life in the ocean \cite{litrevmot:2021:luo}. Additionally, MOT serves as a critical subsystem in autonomous vehicle systems, providing a medium-level representation integral to path-planning processes \cite{selfdriving:2021:badue}.

Detection-Based Tracking is commonly employed in these applications due to its efficacy in scenarios where new objects may emerge or existing objects may exit the frame.
Within a DBT framework, MOT can be represented as a graph optimization problem. In this context, an object detector generates bounding boxes of detected objects in every frame, which serve as graph nodes, and two nodes are connected if and only if they belong to the track of the same object. The basic constraints for this problem enforce three conditions: no backward edge is selected, each node is connected to at most one other node, and no node has multiple incoming nodes.
Thus, the MOT problem is reduced to determining the optimal association between nodes in the graph. Approaches like Tracktor++ \cite{tracktorplusplus:2019:bergmann} perform such graph optimization under the assumption of a high frame rate and available detections in every frame.
However, restricting association to every pair of consecutive frames may be overly simplistic since it barely reflects real-world scenarios where camera motion and occlusion exist.
\begin{figure}[t]
    \centering
    \captionsetup[subfigure]{labelformat=empty, format=plain}
    
    \begin{subfigure}{0.235\textwidth}
        \centering
        \includegraphics[width=\textwidth]{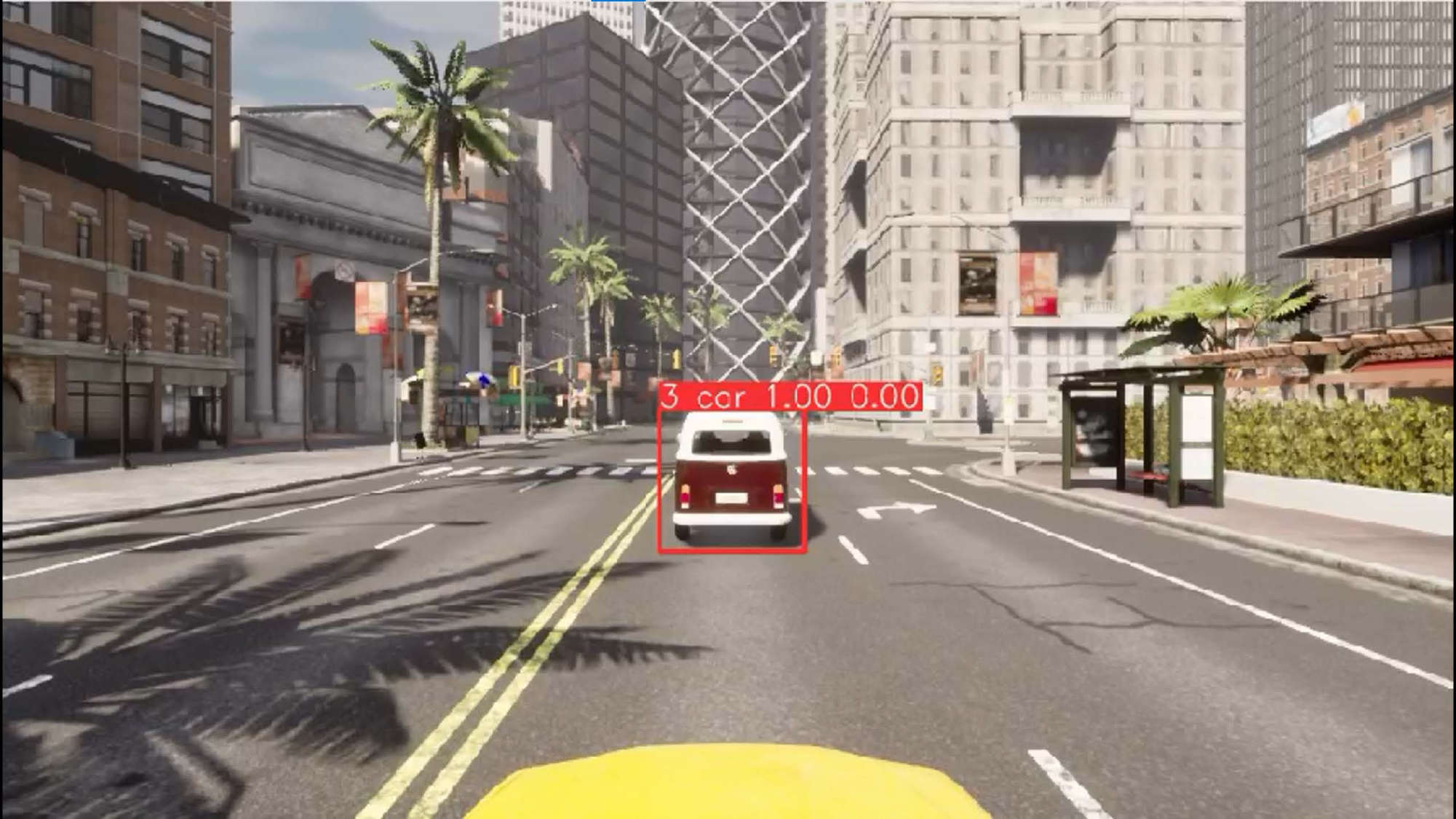}
        \caption{}
    \end{subfigure}\hspace{2pt}
    \begin{subfigure}{0.235\textwidth}
        \centering
        \includegraphics[width=\textwidth]{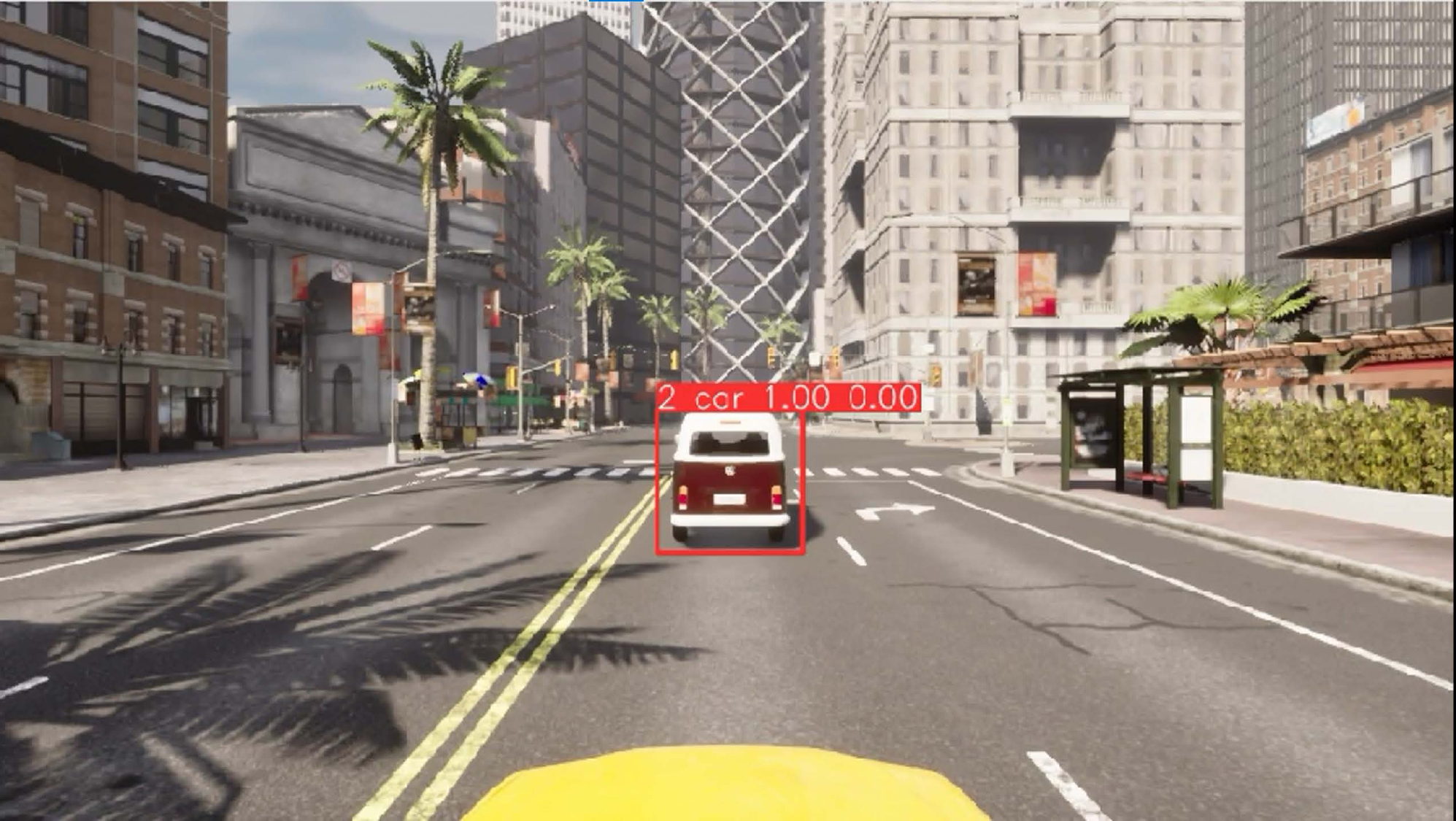}
        \caption{}
    \end{subfigure}
    \\[-4mm]
    \begin{subfigure}{0.235\textwidth}
        \centering
        \includegraphics[width=\textwidth]{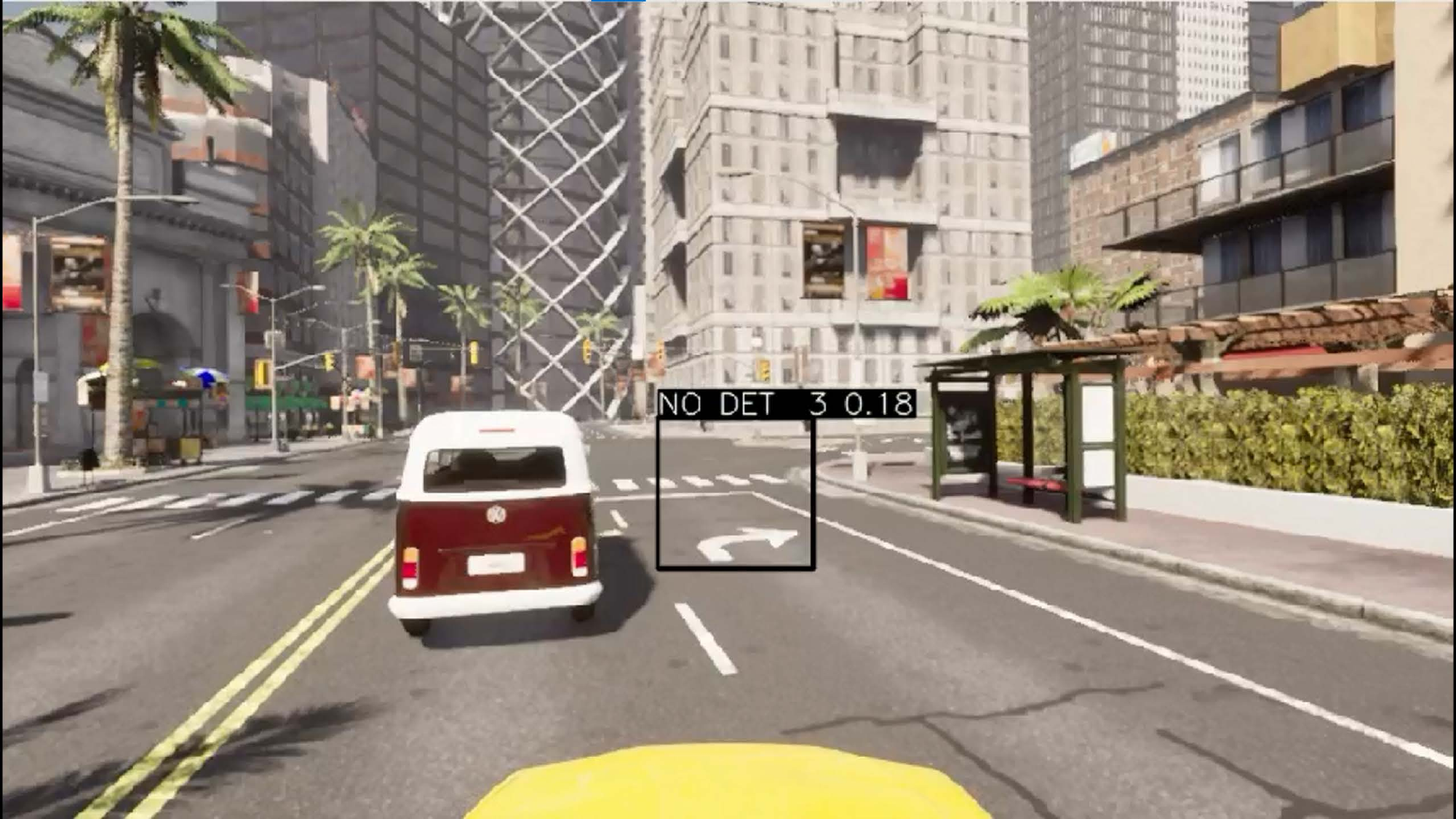}
        \caption{}
    \end{subfigure} \hspace{2pt}%\hfill
    \begin{subfigure}{0.235\textwidth}
        \centering
        \includegraphics[width=\textwidth]{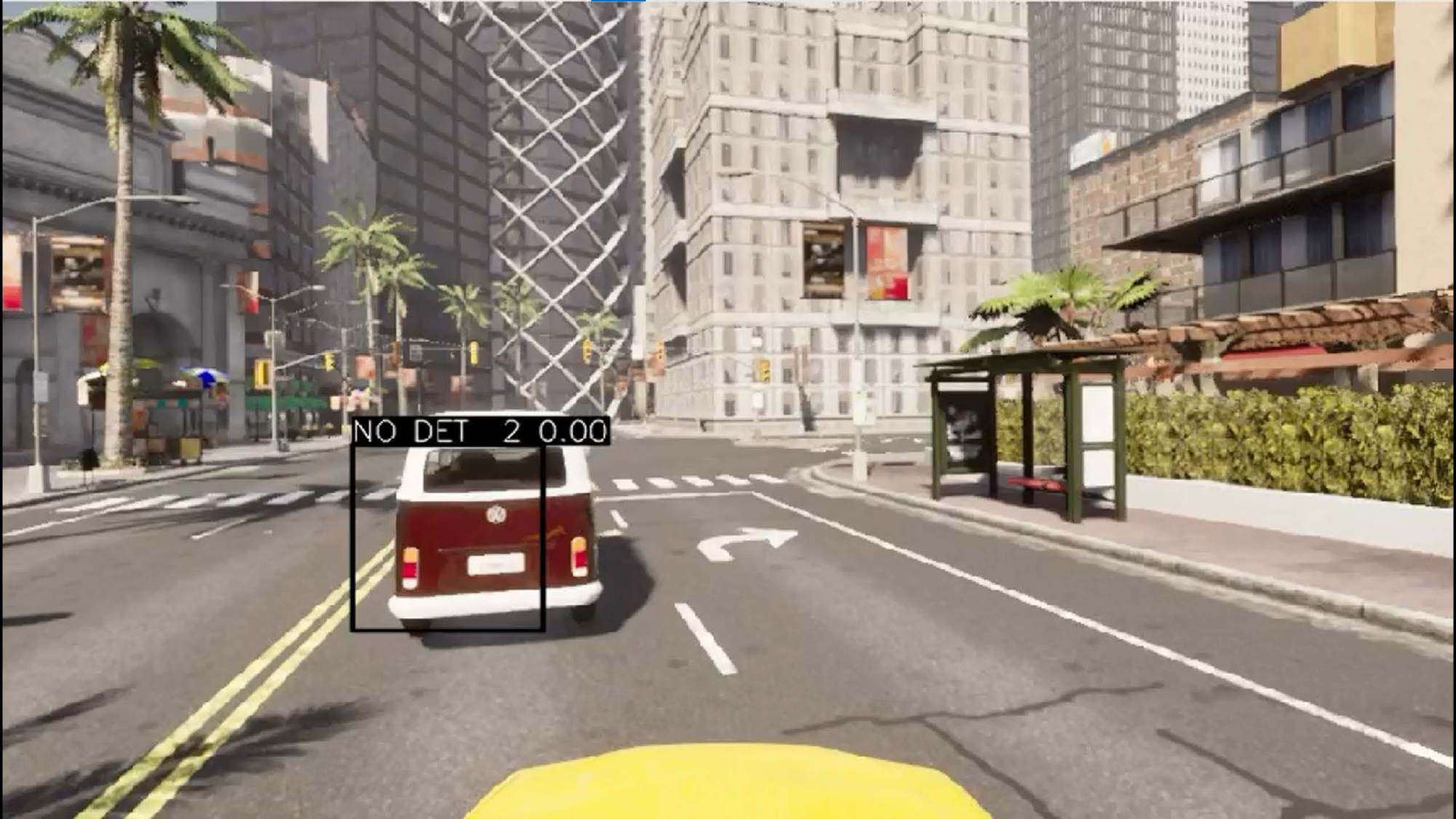}
        \caption{}
    \end{subfigure}
    \\[-4mm]
    \begin{subfigure}{0.235\textwidth}
        \centering
        \includegraphics[width=\textwidth]{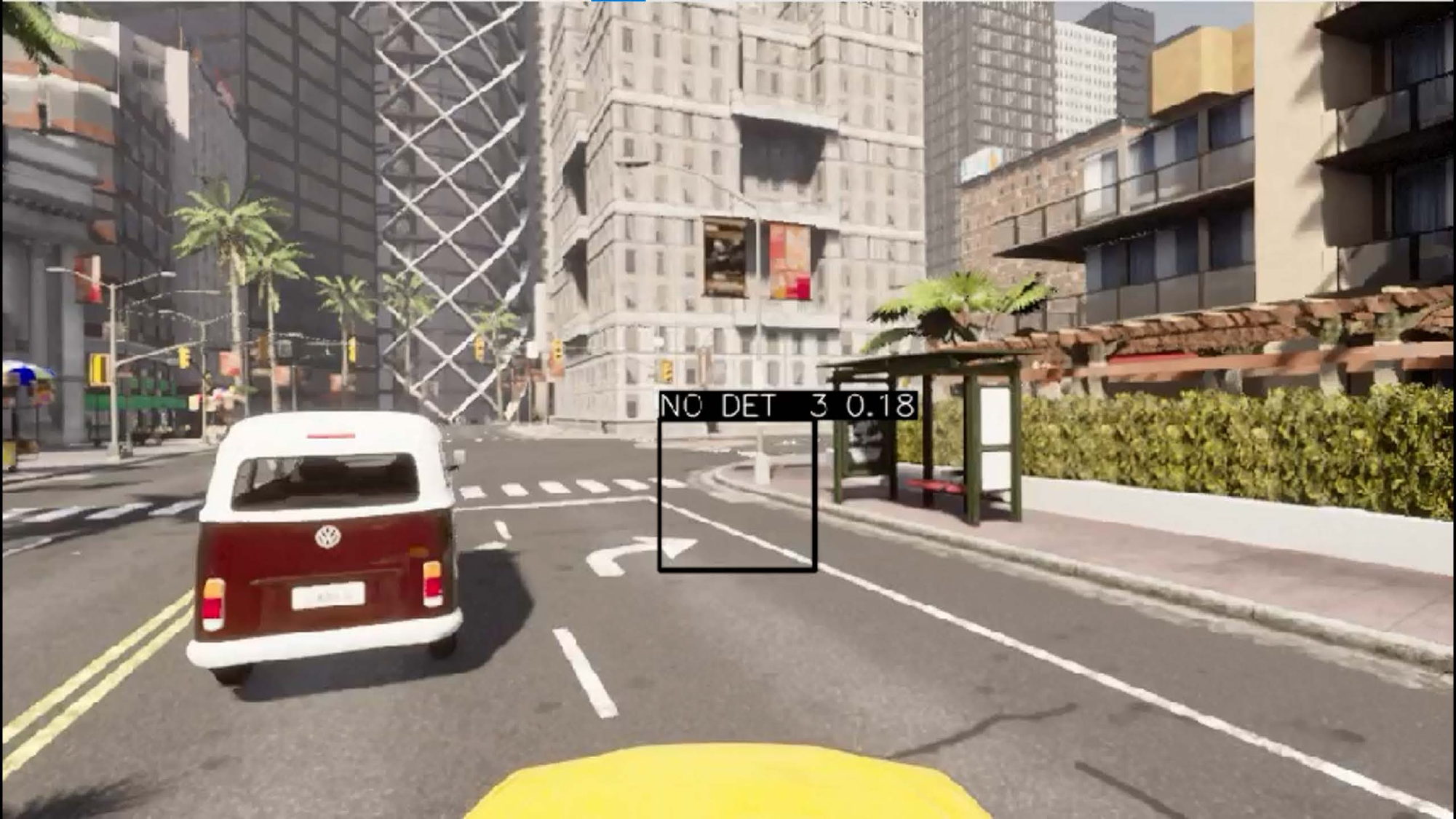}
        \caption{(a)}
    \end{subfigure} \hspace{2pt}%\hfill
    \begin{subfigure}{0.235\textwidth}
        \centering
        \includegraphics[width=\textwidth]{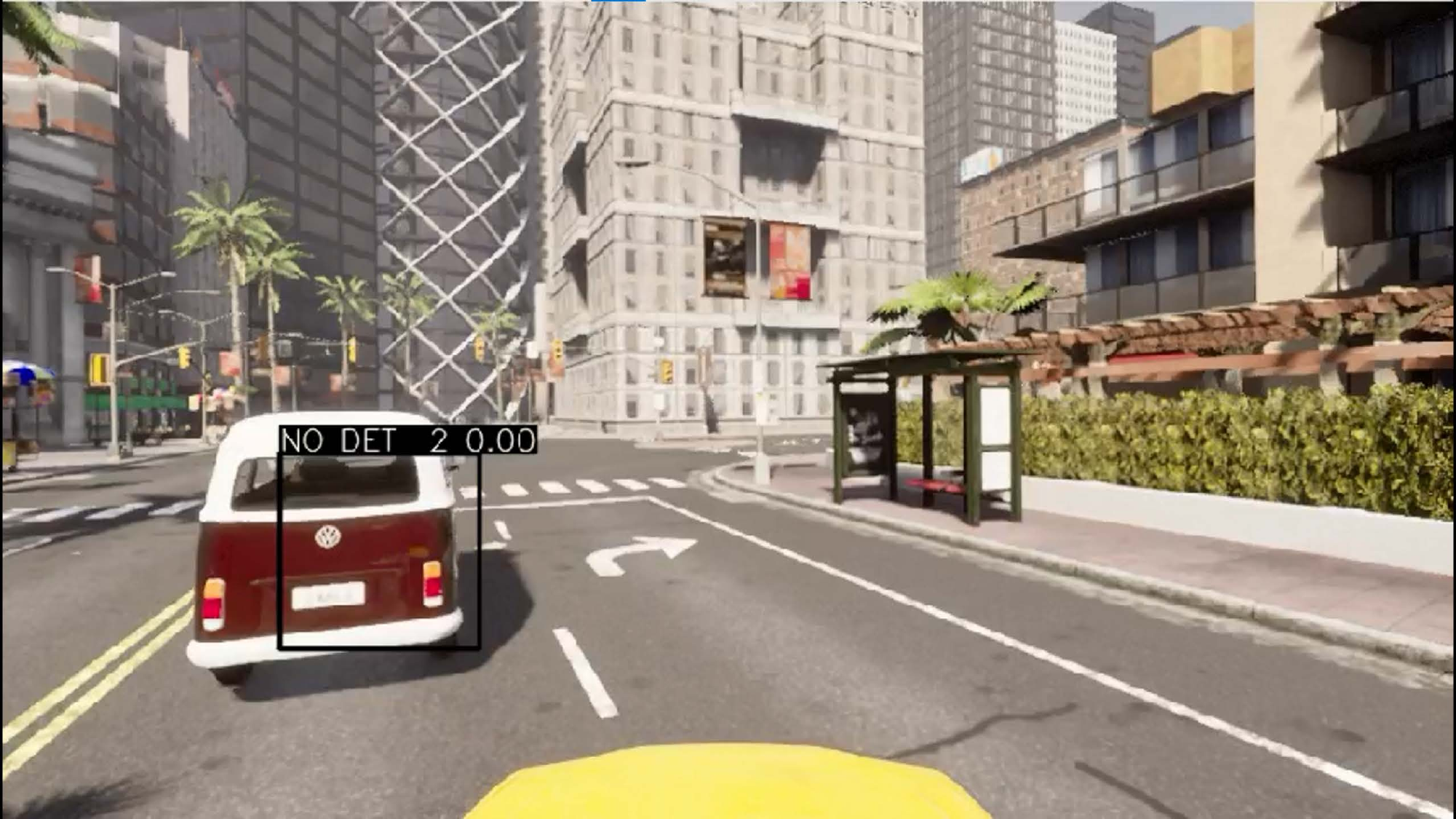}
        \caption{(b)}
    \end{subfigure}

    \caption{In this scenario, the object detection is missed, leading to failure in predicting the location of the object during lane-changing for Vanilla ByteTrack (a). ByteTrack + EMAP (b) significantly improves the prediction, successfully tracking the object in the next two frames.}
    \label{fig:main}
\end{figure}
In situations where detections are not consistently available in every frame, it becomes necessary to either abandon the assumption of association between adjacent frames or substitute the unavailable detection with object state predictions. 
% The illustrative example depicted in Figure \ref{fig:graph-fig} showcases three objects being tracked over four consecutive frames along with their corresponding graph representation. Since the objects are not detected in every frame, the track prediction is employed in lieu of the missing detection.

In KF-based methods, a motion model is assumed for the target objects in the frame and is described by a state transition matrix (\textbf{F}). The rest of the process is modeled using an observation matrix (\textbf{H}), a process noise (\textbf{Q}), an observation noise (\textbf{R}), and an optional disturbance matrix (\textbf{G})
The object location in the image frame is regularly predicted at each time step using the \textit{predict} stage of Kalman Filter:
%% (START) I added this, Feb 21, Mhmd
\begin{equation}
% \begin{aligned}
% &\text{\text{\textit{predict}}} 
\left\{
    \begin{aligned}
    &\hat{\mathbf{x}}_{n+1|n} = \mathbf{F}_n \hat{\mathbf{x}}_{n|n} + [\mathbf{G}_n \mathbf{w}_{n}]\\
    &\mathbf{P}_{n+1|n} = \mathbf{F}_n \mathbf{P}_{n|n} \mathbf{F}_n^\top + \mathbf{Q}_n\\
    \end{aligned}
\right.
\end{equation}
% where \mathbf{F} represents the state transition matrix, and \mathbf{H}, \mathbf{Q}, \mathbf{R} denote the observation matrix H, the process noise Q, and the observation noise
% R, respectively.
Then, newly detected objects are matched with the predicted positions of existing objects, resulting in new observations denoted as \textbf{z}. In the next stage of the Kalman Filter known as the \textit{update} stage, the Kalman gain (\textbf{K}), state estimate ($\hat{\mathbf{x}}$) and estimate covariance (\textbf{P}) are all recalibrated:
\begin{equation}
% &\text{\text{\textit{update}}} 
\left\{
\begin{aligned}
    \mathbf{K}_n &= \mathbf{P}_{n|n-1} \mathbf{H}^\top (\mathbf{H}\mathbf{P}_{n|n-1}\mathbf{H}^\top +\mathbf{R}_n)^{-1}\\
    \hat{\mathbf{x}}_{n|n} &= \hat{\mathbf{x}}_{n|n-1} + \mathbf{K}_n (\mathbf{z}_n - \mathbf{H} \hat{\mathbf{x}}_{n|n-1})\\
    \mathbf{P}_{n|n} &= (\mathbf{I}-\mathbf{K}_n \mathbf{H}) \mathbf{P}_{n|n-1}
\end{aligned}
\right.
\label{eq:predict_and_update}
% \end{aligned}
\end{equation}

%% (END) I added this, Feb 21, Mhmd

Many KF-based approaches are constructed upon the foundation of SORT \cite{sort:2016:Bewley}, which defines the state for each target as follows:
\begin{equation}
\label{equ:sortequ}
 \mathbf{x}=[u, v, s, r, \dot{u}, \dot{v}, \dot{s}]^\top
\end{equation}
In this state definition, $u$ and $v$ represent the horizontal and vertical pixel locations of the center of the target object, while $s$ and $r$ denote the area and height of the bounding box, respectively. Notably, this definition operates under the assumption of a constant velocity model for the tracked objects.
While such an assumption may not hold for all objects, it serves as a reasonable estimation in scenarios where the camera is stationary or has minimal movements, as exemplified by the MOT20 dataset \cite{mot20:2020:dendorfer}.
In scenarios where the camera is mounted on a moving vehicle, the impact of ego-vehicle movement as well as the object motion invalidates the target's constant velocity motion. Specifically, as the motion of the camera becomes more dynamic, the accuracy of predicted \(\dot{u}\) and \(\dot{v}\) diminishes, even when the tracked object maintains a constant velocity.
In Figure \ref{fig:main} (a), the performance of the Kalman Filter is compromised due to rapid camera motion. This failure becomes particularly pronounced in scenarios such as lane changing and taking turns, where the constant-velocity Kalman Filter struggles to make accurate predictions when new observations are unavailable to compensate for the changing dynamics.

\begin{figure}[t]
    \centering
    \includegraphics[width=0.50\textwidth]{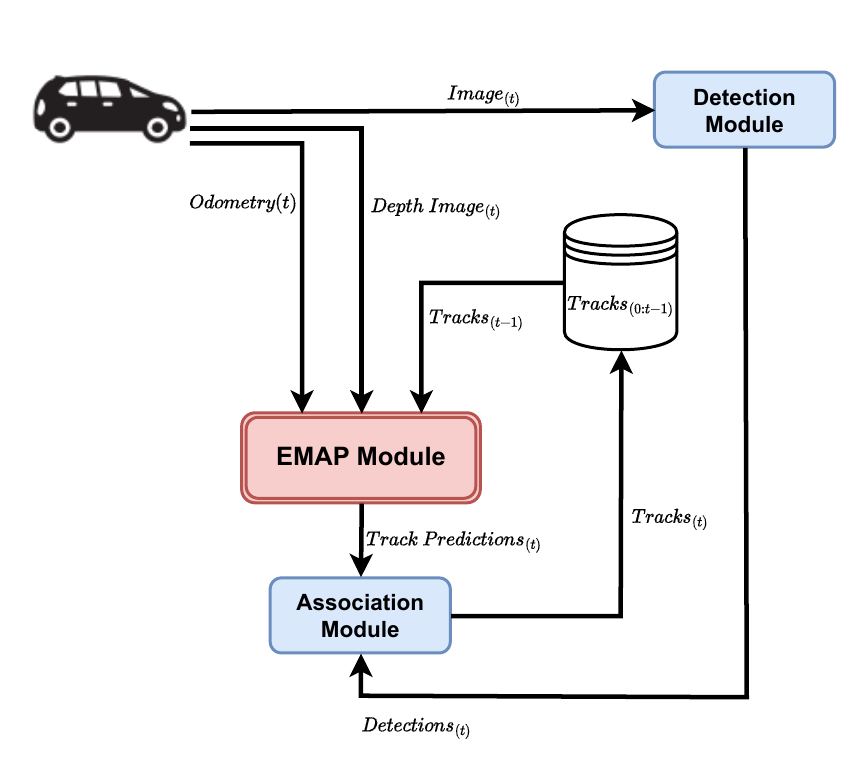}
    \caption{Diagram illustrating the three phases of a detection-based multi-object tracking algorithm. The figure highlights the integration of the EMAP module within the system.}
    \label{fig:maindiagram}
\end{figure}

In this paper, we propose the Ego-Motion Aware Target Prediction module and its integration into KF-based methods. This module incorporates the motion of the ego-vehicle into Kalman Filter, thereby enhancing the performance of MOT solutions.
Our contributions can be summarized as follows:
\begin{itemize}
% \item We introduce the EMAP module designed to enhance the performance of detection-based MOT algorithms.
\item We redefine the Kalman Filter state to decouple camera motion from object trajectories, thereby rejecting disturbances caused by camera motion and maximizing the reliability of the object motion model.

\item We demonstrate the power of EMAP module through our comprehensive experimental evaluations on the KITTI dataset \cite{kitti:2012:geiger} as well as our tailored CARLA \cite{carla:2017:dosovitskiy} autonomous driving dataset and report a substantial improvement in tracking performance, particularly manifested by the reduction in the number of identity switches.
% to show how EMAP can enhance MOT robustness, particularly manifested by the reduction in the number of IDSW.
% when EMAP is integrated into the state-of-the-art MOT algorithms.
% Notably, the heuristic nature of EMAP ensures this enhancement comes with negligible added computational overhead.
\end{itemize}

\section{Related Work}
The related work is studied through two parallel approaches: \textit{Detection-based Tracking}, and \textit{Motion Modeling}.
\paragraph{Detection-based Tracking}

Most of the leading MOT approaches \cite{sort:2016:Bewley, bytetrack:2022:zhang, ocsort:2023:cao, strongsort:2023:du, deepsort:2023:wojke, mat:2022:han}  follow the detection-based tracking schema which consists of three distinct consecutive tasks, namely, detection, prediction, and association.
DBT methods have gained significant attention, driven by notable advancements in the reliability of object detectors in recent years  \cite{yolov8:2023:jocher, efficientdet:2020:tan, retinanet:2017:lin}, enabling DBT techniques to primarily focus on finding the optimal association between per-frame detection bounding boxes. 
% DBT algorithms are also referred to as graph model-based tracking as the problem can be redefined as a graph optimization problem.
% While this provides the advantage of modularity, the overhead of each task is also considerable. 
However, the overall performance of such techniques is highly dependent on the detection module. As a result, when the detector fails to accurately detect target objects, the prediction module is responsible for predicting the imminent location of the missed object until either the object is detected again or 
its track is terminated permanently. In the prediction part, the goal is to find the best prediction of the object location in the next frames based on the previous trajectory of the target. Many dominant methods \cite{ocsort:2023:cao, strongsort:2023:du, deepsort:2023:wojke} use Kalman Filter as their prediction model.
In \cite{sort:2016:Bewley}, the authors exploited Kalman Filter with a constant velocity motion model for the targets to anticipate their future location.
Building on their work, OC-SORT \cite{ocsort:2023:cao} expanded upon the same concept and proposed a method to mitigate the accumulated noise during occlusions. This is achieved by generating a virtual trajectory for the occlusion period while maintaining the simplicity of the Kalman filter model.

Some methods \cite{untrackable:2017:sadeghian, rnnmot:2017:milan} view the prediction and association phase as a joint problem and try to solve the problem using RNNs. For instance, the authors of \cite{untrackable:2017:sadeghian} presented a structure of RNNs that uses motion, appearance, and interaction features of the objects to perform prediction and association based on the cues over a temporal window. Several works \cite{deepsort:2023:wojke, gmphd:2021:baisa, unsreid:2022:liu} concentrated on enriching the association phase by utilizing the appearance features. In \cite{deepsort:2023:wojke} the authors proposed a deep association metric by training a CNN on MARS \cite{mars:2016:zheng} person reidentification dataset.
\cite{gmphd:2021:baisa} designed an identification CNN network to calculate the visual resemblance among intra-frame target bounding boxes and a Gaussian mixture probability hypothesis density filter \cite{gmphd:2006:vo} to track multiple targets across varying densities of targets.
\cite{unsreid:2022:liu} introduced an innovative unsupervised re-identification learning module, eliminating the need for object labels. It incorporated an occlusion estimation module to predict occluded areas, achieving results on par with supervised approaches.

\paragraph{Motion Modeling}

SORT-based tracking approaches leverage a constant velocity model to represent and predict the location of targets within the frame. While this simple model can accurately represent the target movement in occlusion-free static-camera environments, its generalizability decreases in moving camera scenarios specifically when occlusion occurs. Several trackers \cite{mat:2022:han, exploit:2019:wang, mmot:2022:liu, siammot:2021:shuai, delving:2023:huang} used motion cues from objects and the ego-camera to provide a sturdier motion model. MAT \cite{mat:2022:han} utilized the Enhanced Correlation Coefficient Maximization (ECC) \cite{ecc:2008:evangelidis} model to estimate the rotation and translational motion of the camera. It also employs Kalman Filter to predict the location of the targets. \cite{delving:2023:huang} introduced a motion-aware matching module to match tracks with the new observations based on the motion features. 

Nevertheless, while these motion modeling approaches utilized motion clues to increase the reliability of MOT, none of them directly applied the motion information to the Kalman Filter representation. In this work, we adopt the SORT method and integrate ego-motion data into the state definition and Kalman Filter formulation to reject the disturbances and increase the reliability.

\begin{algorithm}[htbp]
\small
\caption{Integrating EMAP with a sample base tracker. \textcolor{codegreen}{Green} highlights our main contribution. }
\begin{algorithmic}[1]

\renewcommand{\algorithmicrequire}{\textbf{Input:}}
\renewcommand{\algorithmicensure}{\textbf{Output:}}
\label{alg:modcode}

\REQUIRE Existing tracks buffer $\mathbb{T}$, New Detections $\mathbb{D}$, Camera translational and rotational motion ($\dot{D}$, $\dot{\psi}$), Depth image $I_d$
\vspace{1mm}
\ENSURE
Updated tracks buffer $\mathbb{T}^{\mathrm{updated}}$
\vspace{1mm}
\STATE Initialize $\mathbb{T}^{\mathrm{pred}} \leftarrow \varnothing$
\vspace{1mm}
\FOR{track $t$ in $\mathbb{T}$}
    \IF{\texttt{RotationOnly}}
        \vspace{1mm}
        \STATE \color{codegreen}{$\mathbb{T}^{\mathrm{pred}} \leftarrow \mathbb{T}^{\mathrm{pred}} \, \cup$ \texttt{Pred($t$, $\dot{\psi}$)}}
        \vspace{1mm}
    \ELSIF{\texttt{TranslationOnly}}
        \vspace{1mm}
        \STATE \color{codegreen}{$\mathbb{T}^{\mathrm{pred}} \leftarrow \mathbb{T}^{\mathrm{pred}} \, \cup$ \texttt{Pred($t$, $\dot{D}$, $I_d$)}}
        \vspace{1mm}
    \ELSE
        \vspace{1mm}
        \STATE \color{codegreen}{$\mathbb{T}^{\mathrm{pred}} \leftarrow \mathbb{T}^{\mathrm{pred}} \, \cup$ \texttt{Pred($t$, $\dot{D}$, $\dot{\psi}$, $I_d$)}}
        \vspace{1mm}
    \ENDIF   
\ENDFOR
\vspace{1mm}
\STATE ($\mathbb{T}^{\mathrm{matched}}$, $\mathbb{D}^{\mathrm{matched}}$) $\leftarrow$ \texttt{Associate($\mathbb{T}^{\mathrm{pred}}, \mathbb{D}$)}
\vspace{1mm}
\STATE $\mathbb{T}^{\mathrm{unmatched}}$ $\leftarrow$ $\mathbb{T} \setminus \mathbb{T}^{\mathrm{matched}}$
\vspace{1mm}
\STATE $\mathbb{D}^{\mathrm{unmatched}}$ $\leftarrow$ $\mathbb{D} \setminus \mathbb{D}^{\mathrm{matched}}$
\vspace{1mm}
\STATE \texttt{UpdateKFstates($\mathbb{T}^{\mathrm{matched}}$, $\mathbb{D}^{\mathrm{matched}}$)}
\vspace{1mm}
\STATE Identify $\mathbb{T}^{\mathrm{lost}}$ from the $\mathbb{T}^{\mathrm{unmatched}}$
\vspace{1mm}
\STATE Create $\mathbb{T}^{\mathrm{new}}$ from the $\mathbb{D}^{\mathrm{unmatched}}$
\vspace{1mm}
\STATE $\mathbb{T}^{\mathrm{updated}} \leftarrow \mathbb{T}^{\mathrm{matched}} \, \cup \, \mathbb{T}^{\mathrm{new}} \cup \, \{ \mathbb{T}^{\mathrm{unmatched}} \setminus  \mathbb{T}^{\mathrm{lost}} \}$

\end{algorithmic}

\end{algorithm}

\section{Methodology}
In the following, the derivation of EMAP, including the projection, integration with Kalman Filter, and isolation of camera motion is discussed in detail.
\subsection{Camera motion projection}
Assuming a rectified image is at hand and using the pinhole camera model, the translational movement of the ego-vehicle affects the target object's location on the image frame according to the following equations:
\begin{equation}
\hat{u}_{trans} = f\frac{d\sin(\frac{u}{f})}{d\cos(\frac{u}{f})-\Delta t\dot{D}}
\end{equation}
\begin{equation}
\hat{v}_{trans} = f\frac{d\sin(\frac{v}{f})}{d\cos(\frac{v}{f})-\Delta t \dot{D}}
\end{equation}
where $f$ is the camera focal length, and $d$ is the Euclidean distance between the object and the vehicle. Also, $\dot{D}$ denotes the ego-vehicle forward translational velocity toward the normal vector of the image plane; $u$ and $v$ are the horizontal and vertical pixel location of the target from the center of image, and $\hat{u}, \hat{v}$ are their respective predictions. 

The impact of ego-vehicle rotation on the pixel location of an object is purely horizontal and governed by:
\begin{equation}
\hat{u}_{rot} = f \frac{\tan(\Delta t \dot{\psi}) + \frac{u}{f}}{1- \frac{u}{f}\tan(\Delta t \dot{\psi})}
\end{equation}
where $\dot{\psi}$ is the ego-vehicle yaw rate. Assuming a small time difference between each two consecutive frames leads to linearized formulations of the form:
\begin{equation}
\hat{u}_{trans} \approx \frac{u\sqrt{u^2+f^2}}{f d} \cdot \dot{D} \Delta t  + u
\end{equation}
\begin{equation}
\hat{v}_{trans} \approx \frac{v\sqrt{v^2+f^2}}{f d} \cdot \dot{D} \Delta t + v
\end{equation}
\begin{equation}
\hat{u}_{rot} \approx f\left(1 + \frac{u^2}{f^2} \right) \cdot \dot{\psi} \Delta t+ u
\end{equation}

\begin{figure}[ht]
    \centering
    \includegraphics[width=0.70\linewidth]{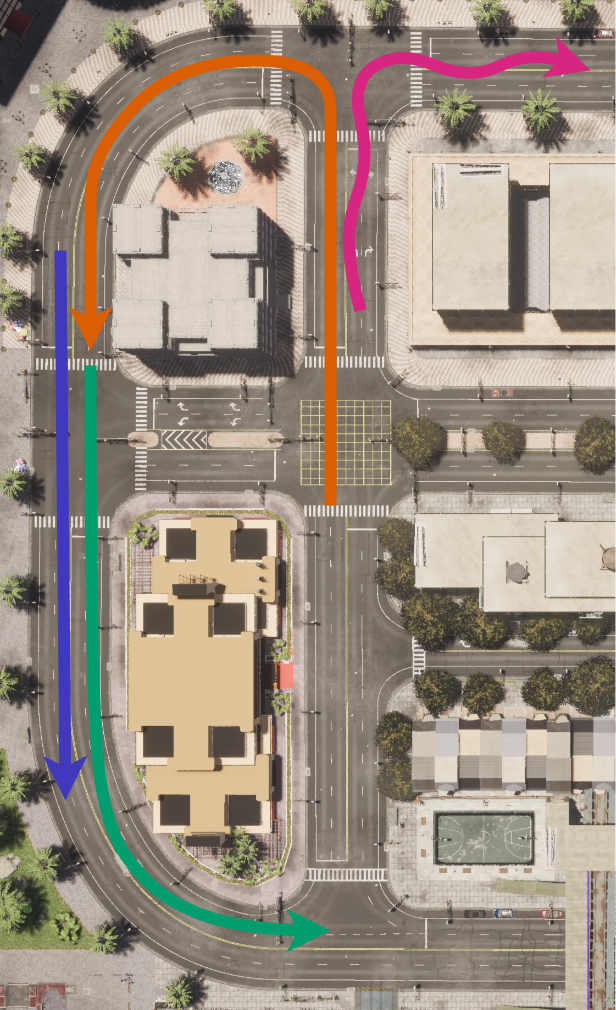}
    \caption{Visualization of \textit{town \#10} in CARLA simulator, featuring four distinct simulation scenarios superimposed on the map. The paths illustrate the diverse trajectories taken in our dataset, capturing a range of scenarios for comprehensive analysis.}
    \label{fig:carla}
    \vspace{-6mm}
\end{figure}

\subsection{Camera motion integration with Kalman filter}
Considering the availability of scene depth map and camera motion information, we can now reformulate the Kalman Filter state definition to integrate this information into a base model. 
% SORT-based methods approximate each object's motion independently by a linear constant velocity model considering eight states for each object as
% $$ \mathbf{x}=[u, v, s, r, \dot{u}, \dot{v}, \dot{s}]^T.$$ 
% In this state definition, $u$ and $v$ show the horizontal and vertical pixel location of the center of the target object while $s$ and $r$ represent the area and the height of the bounding box.
Following the similar format used by SORT in equation \ref{equ:sortequ}, we approximate the motion of the object by a constant velocity model; however, we decouple the effect of camera motion in this motion model. 
% target motion in frame = target motion due to object motion + target motion due to camera movement

Each target object is then modeled using the following state definition:
\begin{equation}
\mathbf{x}=[u^l, v^t, u^r, v^b, {\dot{x}}^l, {\dot{y}}^t, {\dot{x}}^r, {\dot{y}}^b]^\top
\end{equation}
Where $(u^l, v^t)$ indicates the top-left corner of a bounding box, while $(u^r, v^b)$ represents the bottom-right corner of that. Unlike SORT which uses derivatives of $(u, v)$ as noted in \ref{equ:sortequ}, we use ${\dot{x}}^l, {\dot{y}}^t, {\dot{x}}^r$, and ${\dot{y}}^b $. These variables denote the projected horizontal and vertical pixel velocities of the corners of the target object bounding box caused solely by the object's motion in the world frame.
Assuming that the movement of the ego-vehicle can be approximated by two independent parameters, i.e., orientation and forward displacement, we now define the state equation of Kalman Filter as follows:

\begin{equation}
\mathbf{\hat{x}_{n+1}} = \begin{bmatrix}
I_{4\times4} & dtI_{4\times4}\\
0_{4\times4} & I_{4\times4}
\end{bmatrix} \mathbf{\hat{x}_{n}} + \begin{bmatrix}
    G_{n}^\psi & G_{n}^D\\
\end{bmatrix} \begin{bmatrix}
    \dot{\psi} \\
    \dot{D}
\end{bmatrix} dt
\label{state:pred}
\end{equation}
where,
\begin{equation} 
G_{n}^\psi = 
\begin{bmatrix}
    f\left(1 + \left(\frac{u_{n}^l}{f}\right)^2 \right) \\ \\
    0 \\ \\
    f\left(1 + \left(\frac{u_{n}^r}{f}\right)^2 \right) \\ \\
    0\\ \\
    0_{4\times1}
\end{bmatrix}, \quad
% \end{equation}
% and
% \begin{equation} 
G_{n}^D = 
\begin{bmatrix}
    \frac{u_{n}^l\sqrt{(u_{n}^l)^2+f^2}}{f d_n} \\ \\
    \frac{v_{n}^t\sqrt{(v_{n}^t)^2+f^2}}{f d_n} \\ \\
    \frac{u_{n}^l\sqrt{(u_{n}^l)^2+f^2}}{f d_n} \\ \\
    \frac{v_{n}^b\sqrt{(v_{n}^b)^2+f^2}}{f d_n} \\ \\
    0_{4\times1}
\end{bmatrix}
\end{equation}
These equations describe the \textcolor{codegreen}{\texttt{Pred}} function in Algorithm \ref{alg:modcode}.

When an associated detection of an object exists, Kalman Filter \textit{update} phase is used to re-calibrate the state variables. Then, the \textit{prediction} is performed to extrapolate the location of the target in the next frames. 
% In time steps without any associated detection, we follow the \textit{dummy update} process proposed by \cite{ocsort:2023:cao}, in which the prediction of the state is substituted as the current state of the object.
Figure \ref{fig:maindiagram} shows an overview of the EMAP module integrated into a DBT multi-object tracking framework. 
In the depicted figure, the input image, captured by the camera, is directed into the detection module. 
Simultaneously, the synchronized odometry data and depth image are fed into the EMAP module. Following the generation of predictions by EMAP for the subsequent location of each existing track, the tracks are then associated with the newly detected objects. Finally, the iteration ends when the tracks are updated with their new associated detection or discarded according to the baseline design.
The pseudocode presented in Algorithm \ref{alg:modcode} outlines the integration of our prediction module with existing DBT algorithms. This integration requires two additional inputs, camera motion information, and the depth map.

\subsection{Justification for isolating the camera motion}
As mentioned in section \ref{intro}, most Kalman Filter-based DBT algorithms approximate the motion of the target by constant velocity models independent of other objects and camera motion. 
% However, this assumption can easily be violated in scenarios where camera movements introduce large disturbances to the bounding box location.
% This issue is especially worse for a constant velocity model as it fails to accurately predict the track of the target in the presence of detection gaps.
However, this assumption can be readily violated in scenarios where camera movements introduce significant disturbances to the bounding box location. This issue is particularly exacerbated for a constant velocity model, as it struggles to accurately predict the target's trajectory in the presence of detection gaps.
Such inaccurate predictions give rise to frequent identity switches, ultimately compromising the overall MOT performance.
To address this problem, in EMAP module, we take the camera motion into account isolated from the object motion model. 

\begin{figure}[t]
    \centering
    \includegraphics[width=\linewidth]{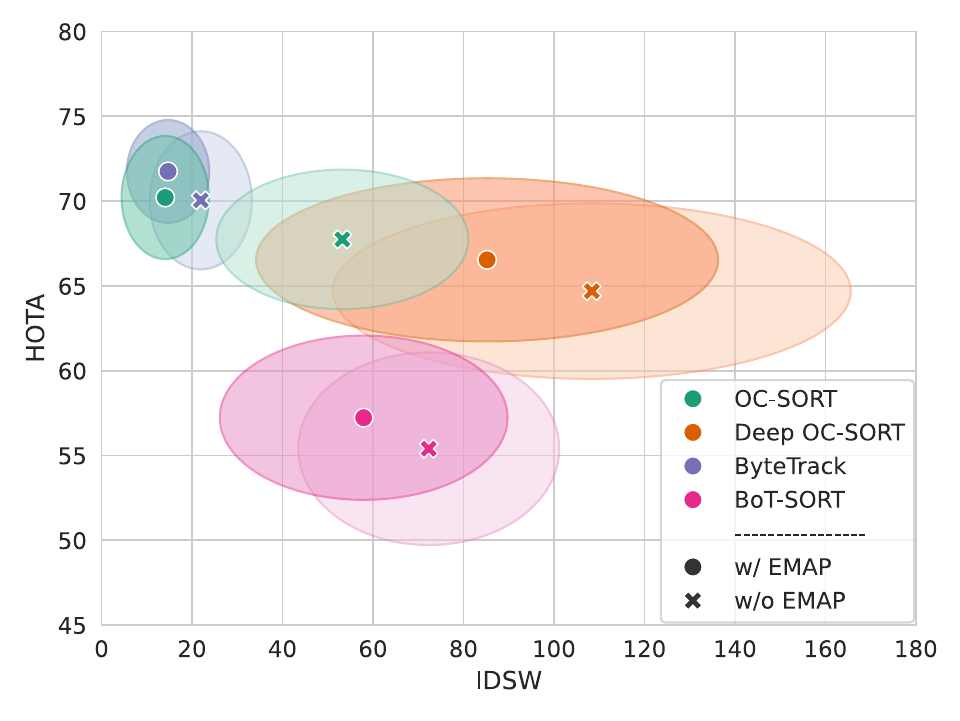}
    \caption{HOTA vs IDSW comparisons of OC-SORT, Deep OC-SORT, ByteTrack, and BoT-SORT on 21 KITTI train sequences with or without EMAP module. The height and width of the ellipses are the standard deviation of the distribution.}
    \label{fig:eval_dist}
\end{figure}
\section{Experiments}
In the following section, the datasets, metrics, baseline trackers, and the results are presented.
\subsection{Datasets}
\subsubsection{KITTI}
The KITTI dataset serves as a cornerstone in benchmarking computer vision algorithms for autonomous driving tasks. It consists of real-world data captured from a car equipped with sensors such as LiDAR, camera, and GPS. In our experiments, we use the KITTI training dataset which consists of 21 sequences of diverse driving scenarios such as urban, highway, and rural environments. For the evaluation, we focus on the \textit{Car} and \textit{Pedestrian} classes for which we take bounding boxes detected by PermaTrack \cite{perma:2021:takmakov} and also YOLOv8 \cite{yolo:2016:redmon}.

\subsubsection{CARLA Simulation Dataset}
In addition to the KITTI dataset, we conducted experiments using a custom-generated dataset within the CARLA autonomous driving simulator. The CARLA simulator offers a realistic virtual environment for autonomous driving, allowing for controlled and diverse scenarios. In our dataset, we have four sequences each representing distinct features. In \textcolor{myblue}{Scenario \#1}, a car navigates a straight road. \textcolor{mygreen}{Scenario \#2} involves the car initiating movement from a stationary position, moving along a straight path, encountering a curve, and concluding the sequence upon completing the curve. \textcolor{myorange}{Scenario \#3} features the car starting from a stationary position, progressing along a straight trajectory, taking a left turn, reaching a curve, and coming to a stop afterward. In \textcolor{mymagenta}{Scenario \#4}, the vehicle travels a path with winding movements ending after taking a right turn. Figure \ref{fig:carla} illustrates the paths on the CARLA \textit{town \#10} map. We integrate CARLA with MOT algorithms through the Robot Operating System (ROS) \cite{ros:2009:quigley}, enabling communication between Python and CARLA. Synchronized messages, including RGB images, depth map (aligned with RGB), ego-vehicle odometry, and automatically generated ground truth tracking bounding boxes from CARLA are streamed to Python via ROS for seamless evaluation.

\subsection{Evaluation Metrics}
We select HOTA \cite{hota:2021:luiten} as the primary metric for its ability to balance between the accuracy of both object detection and association in multi-object tracking. Also, we report IDSW as the additional primary metric because it represents the reliability of generated trajectories. The underlying reason is that IDSW counts the occurrences of incorrect object identity swaps or the loss and subsequent reinitialization of identities.
Our analysis extends to Association Accuracy (AssA) for evaluating association performance, along with IDF1 as an additional metric in the same context. It is worth mentioning that MOTA, which used to be the predominant measure of MOT performance, is heavily affected by the detection quality. Hence, to ensure a fair comparison, we use identical detections among all algorithms to report the MOTA score.

\subsection{Baseline SORT-based Algorithms}
In order to evaluate the impact of EMAP on existing MOT algorithms, we selected four state-of-the-art SORT-based algorithms and integrated EMAP with their prediction module to evaluate the effect of EMAP on their performance. The algorithms we selected as baseline are OC-SORT \cite{ocsort:2023:cao}, Deep OC-SORT \cite{deepocsort:2023:maggiolino}, ByteTrack \cite{bytetrack:2022:zhang}, and BoT-SORT \cite{botsort:2022:aharon}. All four baselines are Kalman Filter-based algorithms. However, in BoT-SORT, authors use image registration to approximate the camera motion and compensate for the effect of this motion.

% \begin{figure*}[t]
%     \centering
%     \captionsetup[subfigure]{labelformat=empty, format=plain}
    
%     \begin{subfigure}{0.485\textwidth}
%         \centering
%         \includegraphics[width=\textwidth]{figures/exp-compare/hota.pdf}
%         \caption{}
%     \end{subfigure}\hfill
%     \begin{subfigure}{0.485\textwidth}
%         \centering
%         \includegraphics[width=\textwidth]{figures/exp-compare/idsw.pdf}
%         \caption{}
%     \end{subfigure}
    
% \caption{scscsdvdvvsdv.}
% \label{fig:errorbox}
% \end{figure*}

\subsection{Results}

In this section, we present the performance of our solution on both the CARLA and the KITTI datasets.
% Table \ref{table:simresyolo} showcases the metrics evaluated on the simulation dataset with YOLOv8x as the object detector. It demonstrates that incorporating EMAP enhances the performance of OC-SORT and ByteTrack, resulting in improved HOTA scores and reduced identity switches. However, Deep OC-SORT experiences a slight decline in performance, while BOT-SORT exhibits a significant drop in HOTA. This decline in the performance of BOT-SORT can be attributed to the doubling effect of camera motion compensation by EMAP, leading to erroneous predictions.
Table \ref{table:resbyseq} displays the HOTA scores and identity switch counts for each sequence in the CARLA simulation dataset. Notably, EMAP significantly impacts Sequence \#4, characterized by substantial vehicle rotational and translational movements. This sequence shows a decrease in identity switches and an increase in HOTA values for all trackers, highlighting EMAP's effectiveness in complex motion scenarios.

Table \ref{table:kittiresperma} presents the average metric values across the 21 sequences of the KITTI train dataset when detections are taken from PermaTrack. EMAP notably reduces the number of identity switches across all four baselines, with the most significant impact observed in OC-SORT, where it decreases by 76\%. Furthermore, EMAP integration improves various metrics such as HOTA, MOTA, IDF1, FP, FN, AssA, and AssR in all four baseline trackers. Figure \ref{fig:eval_dist} illustrates that integration of the EMAP module notably decreases identity switches and enhances HOTA scores compared to baseline performance. 
Additionally, it demonstrates that EMAP contributes to the improved robustness of baseline trackers, as evidenced by the reduced variance in IDSW and HOTA across all sequences.

In Table \ref{table:kittiresyolo}, we present the evaluation on KITTI train dataset using YOLOv8x as the object detector. % to illustrate the results with online object detection.
Despite the fact that YOLOv8x exhibits weaker performance compared to PermaTrack indicated by higher missed detections and lower detection accuracy, 
the addition of EMAP still positively affects the performance of all baseline trackers.%, albeit to a lesser extent than with PermaTrack.

% (To further showcasing the power of EMAP, we select the sequences with high ego-vehicle movements including both rotational and translational motions)........

To assess the individual and combined impact of the translation and rotational submodules, we conduct an ablation study comparing the results of EMAP with each submodule added separately and together, as shown in Table \ref{table:res_abl}. The largest reduction in the number of identity switches occurs when both submodules are active. However, the maximum HOTA is attained when only the rotational module is activated.

\begin{table*}[hbt!]
\centering
\caption{Performance comparison of MOT algorithms on our CARLA dataset split by sequence with YOLOv8x as the object detector. The best results are shown in \textbf{bold}.}
\setlength{\tabcolsep}{7pt}
\scriptsize
\begin{tabular}{l p{20px} p{20px} p{20px} p{20px} p{20px} p{20px} p{20px} p{20px}}
\toprule
\multirow{2}{*}{Tracker} & \multicolumn{2}{c}{Sequence \#1} & \multicolumn{2}{c}{Sequence \#2} & \multicolumn{2}{c}{Sequence \#3} & \multicolumn{2}{c}{Sequence \#4}\\ \cmidrule(lr){2-3} \cmidrule(lr){4-5} \cmidrule(lr){6-7} \cmidrule(lr){8-9}
 &  HOTA$\uparrow$ & IDs$\downarrow$  &  HOTA$\uparrow$ & IDs$\downarrow$ &  HOTA$\uparrow$ & IDs$\downarrow$ &  HOTA$\uparrow$ & IDs$\downarrow$\\
\midrule
OC-SORT~\cite{ocsort:2023:cao} &36.15  & 5& 24.84& 11& 36.3& 15& 24.09& 19\\
OC-SORT + EMAP~\cite{ocsort:2023:cao} &\textbf{36.24}  & \textbf{1}& \textbf{25.06}& \textbf{3}& \textbf{37.04}& 15& \textbf{29.09}& \textbf{7}\\

BoT-SORT~\cite{botsort:2022:aharon} &\textbf{35.08} & 7 & 21.88& 35& \textbf{32.64}& 30& 23.44& 24\\
BoT-SORT + EMAP~\cite{botsort:2022:aharon} &34.94 & 7 & \textbf{22.14}& 35& 31.65& 30& \textbf{24.54}& \textbf{18}\\

Deep OC-SORT~\cite{deepocsort:2023:maggiolino} &\textbf{35.71} &\textbf{4} &\textbf{ 23.65}& \textbf{55}& \textbf{36.33}&\textbf{ 36}& 25.17& 67\\
Deep OC-SORT + EMAP ~\cite{deepocsort:2023:maggiolino} &35.55 &9 & 20.75& 129& 35.55& 61& \textbf{26.66}& \textbf{38}\\

ByteTrack~\cite{bytetrack:2022:zhang} &34.94 & 0& 21.14& 1&34.26 & 5& 25.71& 8\\ 
ByteTrack + EMAP  &\textbf{25.05} & 0& \textbf{21.7}& \textbf{0}& \textbf{34.91}& \textbf{4}& \textbf{28.39}& \textbf{2}\\
\bottomrule
\end{tabular}
\label{table:resbyseq}
\end{table*}

\begin{table*}[hbt!]
\centering
\caption{Average performance (on 21 sequences) on KITTI-train dataset, detections are taken from PermaTrack. The best results are shown in \textbf{bold}.}
\setlength{\tabcolsep}{7pt}
\scriptsize
\begin{tabular}{lrrrrrrrr}
\toprule
Tracker &  HOTA$\uparrow$ & MOTA$\uparrow$ & IDF1$\uparrow$ &  FP$\downarrow$ & FN$\downarrow$ & IDs$\downarrow$ &  AssA$\uparrow$ & AssR$\uparrow$ \\
\midrule
OC-SORT~\cite{ocsort:2023:cao} & 67.74 & 64.53 & 78.24  & 709.33 & 234.19 & 53.24 & 68.38 & 79.34\\
OC-SORT + EMAP& \textbf{70.21} & \textbf{67.87} & \textbf{82.17}  & \textbf{662.19} & \textbf{187.05} & \textbf{14.10} & \textbf{73.11} & \textbf{85.32}\\
\midrule
Deep OC-SORT~\cite{deepocsort:2023:maggiolino} & 64.69 & 62.59 & 74.20  & 893.67 & 418.52 & 108.38 & 62.84 & 74.41 \\
Deep OC-SORT + EMAP & \textbf{66.54} & \textbf{64.38} & \textbf{76.37}  & \textbf{863.14} & \textbf{388.00} & \textbf{85.24} &  \textbf{66.22} & \textbf{77.80} \\
\midrule
ByteTrack~\cite{bytetrack:2022:zhang} & 70.05 & 74.98 & 84.60  & 339.10 & 305.00 & 21.95 &  72.62 & 80.28\\ 
ByteTrack + EMAP & \textbf{71.75} & \textbf{75.82} & \textbf{85.78}  & \textbf{324.33} & \textbf{295.33} & \textbf{14.71} &  \textbf{74.71} & \textbf{81.73} \\
\midrule
BoT-SORT~\cite{botsort:2022:aharon} & 55.40 & 47.74 & 65.89  & 602.95 & 687.29 & 72.33 &  56.36 & 62.33 \\
BoT-SORT + EMAP & \textbf{57.23} &  \textbf{50.11} &  \textbf{68.49} & \textbf{578.29} & \textbf{670.76} &  \textbf{57.95} &   \textbf{59.58} &  \textbf{66.05} \\
\bottomrule
\end{tabular}
\label{table:kittiresperma}
\end{table*}

\begin{table*}[hbt!]
\centering
\caption{Average performance (on 21 sequences) on KITTI-train dataset, detections are taken from YOLOv8. The best results are shown in \textbf{bold}.}
\setlength{\tabcolsep}{7pt}
\scriptsize
\begin{tabular}{lrrrrrrrr}
\toprule
Tracker &  HOTA$\uparrow$ & MOTA$\uparrow$ & IDF1$\uparrow$ &  FP$\downarrow$ & FN$\downarrow$ & IDs$\downarrow$ & AssA$\uparrow$ & AssR$\uparrow$ \\
\midrule
OC-SORT~\cite{ocsort:2023:cao} & 39.35 &  37.0  &  53.04 & 163.14 & 1187.9 &  18.86 &    46.16 &  48.82\\
OC-SORT + EMAP& \textbf{40.28} &  \textbf{38.0}  &  \textbf{54.4}  & 168.05 & \textbf{1179.95}&  \textbf{12.29} &   \textbf{47.68} &  \textbf{50.25}\\
\midrule
Deep OC-SORT~\cite{deepocsort:2023:maggiolino} & 38.22 &  35.77 &  50.08 & 270.67 & 1236.38&  69.38 &   42.21 &  45.36 \\
Deep OC-SORT + EMAP & \textbf{39.6}  &  \textbf{37.2}  &  \textbf{52.88} & \textbf{225.48} & \textbf{1191.19}&  \textbf{41.33} &   \textbf{45.15} &  \textbf{48.41} \\
\midrule
ByteTrack~\cite{bytetrack:2022:zhang} & 33.48 &  31.29 &  44.66 & 102.76 & 1368.76&  14.24 &    42.11 &  44.2\\ 
ByteTrack + EMAP & \textbf{33.78} &  \textbf{31.32} &  \textbf{45.81} &  \textbf{83.95} & \textbf{1356.81}&  \textbf{11.9}  &    \textbf{42.69} &  44.12 \\
\midrule
BoT-SORT~\cite{botsort:2022:aharon} & 25.16 &  23.5  &  34.57 &  138.0 & 1584.1 &  46.05 &    29.17 &  29.81 \\
BoT-SORT + EMAP &  \textbf{28.06} &  \textbf{25.49} &  \textbf{39.21} &  \textbf{123.7} & \textbf{1425.65}&  \textbf{35.2}  &   \textbf{33.82}&  \textbf{34.55} \\
\bottomrule
\end{tabular}
\label{table:kittiresyolo}
\end{table*}

% \begin{table*}[hbt!]
% \centering
% \caption{Average performance of state-of-the-art MOT methods on CARLA dataset with YOLOv8x as the object detector. The best results are shown in \textbf{bold}.}
% \setlength{\tabcolsep}{7pt}
% \scriptsize
% \begin{tabular}{l | rrrrrrrr}
% \toprule
% Tracker &  HOTA$\uparrow$ & MOTA$\uparrow$ & IDF1$\uparrow$ &  FP$\downarrow$ & FN$\downarrow$ & IDs$\downarrow$ &  AssA$\uparrow$ & AssR$\uparrow$ \\
% \midrule
% OC-SORT~\cite{ocsort:2023:cao} & 30.34 &  20.08 &  39.04 & 514.25 & 2367.25&  12.5  &   39.45 &  44.5\\
% OC-SORT + EMAP& 31.86 &  20.21 &  42.29 &  437.5 & 2283.75&   6.5  &    43.28 &  48.22\\
% \midrule
% Deep OC-SORT~\cite{deepocsort:2023:maggiolino} & 30.21 &  18.91 &  37.87 &  572.5 & 2384.0 &  40.5  &   38.71 &  43.62 \\
% Deep OC-SORT + EMAP & 29.63 &  18.34 &  36.35 &  611.5 & 2423.0 &  59.25 &   37.55 &  41.51 \\
% \midrule
% ByteTrack~\cite{bytetrack:2022:zhang} & 29.01 &  20.1  &  37.24 & 327.75 & 2469.0 &   3.5  & 40.66 &  44.75\\ 
% ByteTrack + EMAP & 30.01 &  20.54 &  39.22 & 276.25 & 2419.25&   1.5  &   43.28 &  47.21 \\
% \midrule
% BoT-SORT~\cite{botsort:2022:aharon} & 28.26 &  21.48 &  36.26 & 354.25 & 2490.75&  24.0  &    37.34 &  38.88 \\
% BoT-SORT + EMAP & 25.54 &  16.82 &  31.33 &  458.5 & 2617.25&  39.25 &    33.8  &  35.39 \\
% \bottomrule
% \end{tabular}
% \label{table:simresyolo}
% \end{table*}

\begin{table}[hbt!]
\centering
\caption{Ablation on KITTI dataset with Translational and Rotational sub-modules. The best results are shown in \textbf{bold}.}
\setlength{\tabcolsep}{7pt}
\scriptsize
\begin{tabular}{llrrrrr}
\toprule
Method & Tracker &  HOTA$\uparrow$ & MOTA$\uparrow$ & IDF1$\uparrow$ & IDs$\downarrow$ \\
\midrule
\hfil\multirow[c]{4}{*}{\parbox{34px}{Baseline}} & OC-SORT & 67.74 &  64.53 &  78.24 &  53.24 \\
 & Deep OC-SORT & 64.69 &  62.59 &  74.2  & 108.38 \\
 & ByteTrack & 70.05 &  74.98 &  84.6  &  21.95 \\ 
 & BoT-SORT & 55.4  &  47.74 &  65.89 &  72.33 \\
\midrule
 
\hfil\multirow[c]{4}{*}{\parbox{34px}{Translational-Only EMAP}}  & OC-SORT & 71.45 &  74.59 &  83.35 &  24.38 \\
 & Deep OC-SORT & 65.33 &  63.08 &  74.91 &  99.43 \\
 & ByteTrack & 70.25 &  74.77 &  84.4  &  23.0  \\
 & BoT-SORT  & 55.28 &  48.65 &  65.58 &  76.71 \\

\midrule
\hfil\multirow[c]{4}{*}{\parbox{34px}{Rotational-Only EMAP}}  & OC-SORT & \textbf{72.63} &  75.1  &  85.05 &  15.43  \\
 & Deep OC-SORT & 65.57 &  63.72 &  75.62 &  97.14 \\
 & ByteTrack &  71.18 &  \textbf{75.86} &  85.69 &  15.1  \\
 & BoT-SORT  &55.41 &  48.65 &  65.59 &  85.81 \\

\midrule
\hfil\multirow[c]{4}{*}{\parbox{34px}{EMAP}}  & OC-SORT & 70.21 &  67.87 &  82.17 &  \textbf{14.1}  \\
 & Deep OC-SORT & 66.54 &  64.38 &  76.37 &  85.24 \\
 & ByteTrack & 71.75 &  75.82 &  \textbf{85.78} &  14.71 \\
 & BoT-SORT  & 57.23 &  50.11 &  68.49 &  57.95 \\

\bottomrule
\end{tabular}
\label{table:res_abl}
\end{table}

\subsection{Visualization results}
To showcase the impact of EMAP module, we present a visualization of the Bytetrack MOT algorithm operating within the CARLA simulator under a scenario where detection is lost. As depicted in Figure \ref{fig:main}  (a), the prediction module of the vanilla ByteTrack struggles to accurately anticipate the position of the target after the detection becomes unavailable (denoted by black bounding boxes). In contrast, Figure \ref{fig:main}  (b) illustrates the enhanced performance achieved by incorporating our module. The predicted bounding box location aligns more closely with the actual location, underscoring the efficacy of the ego-motion aware prediction module.

\section{Conclusions and Future Work}

In this paper, we enhance the performance of detection-based MOT algorithms by leveraging both the motion of the camera and depth information.
Our contribution, the Ego-Motion Aware Prediction (EMAP) module, serves as a predictive component seamlessly integrated into DBT multi-object tracking algorithms.
EMAP adeptly rejects disturbances introduced by camera motion, thereby boosting the reliability of the object motion model. Through the reformulation of the Kalman Filter, EMAP effectively decouples the impact of the rotational and translational velocities of the camera from the target location within the image frame.
In our experiments on KITTI train MOT dataset, EMAP significantly enhances the performance of the base trackers, evident in increased HOTA and reduced number of identity switches. Our in-depth investigation in CARLA simulator further reveals the efficacy of EMAP where ego-camera motion dominantly impacts the target location. Our module depicts how the reliability of existing MOT algorithms can be improved in real-world scenarios, especially autonomous driving.
% the CARLA simulator, EMAP significantly elevates the performance of the base tracker, evident in HOTA and notable reductions in the number of identity switches.

As a future direction, we aim to explore the retrieval of camera motion and depth information using Visual Odometry and depth estimation algorithms solely from RGB cameras, with the ultimate goal of extending the versatility of our module to systems relying exclusively on an RGB camera as their primary sensor.

% \section*{Acknowledgment}

\bibliographystyle{IEEEtran}

\bibliography{bib}

\end{document}